\documentclass[letterpaper, twocolumn]{article}

\usepackage{amsmath}
\usepackage{amssymb}
\usepackage{amsthm}
\usepackage{graphicx}
\usepackage{subcaption}
\usepackage{verbatim}
\usepackage[toc,page]{appendix}
\usepackage{color,soul}
\usepackage{breqn}
\usepackage[ruled]{algorithm2e}  
\usepackage{booktabs} 
\usepackage{authblk}
\usepackage{hyperref}
\usepackage{float}

\newcommand\numberthis{\addtocounter{equation}{1}\tag{\theequation}}
\DeclareMathOperator{\EX}{\mathbb{E}}

\author{Daniel Braithwaite\thanks{daniel.braithwaite@ecs.vuw.ac.nz} }
\author{W. Bastiaan Kleijn\thanks{bastiaan.kleijn@ecs.vuw.ac.nz} }
\affil{School of Engineering and Computer Science, Victoria University of Wellington}

\date{}

\begin{document}

\title{Bounded Information Rate Variational Autoencoders}

\maketitle

\begin{abstract}
This paper introduces a new member of the family of Variational Autoencoders (VAE) that
constrains the rate of information transferred by the latent layer. The latent layer is
interpreted as a communication channel, the information rate of which is bound by imposing
a pre-set signal-to-noise ratio. The new constraint subsumes the mutual information
between the input and latent variables, combining naturally with the likelihood objective
of the observed data as used in a conventional VAE. The resulting Bounded-Information-Rate
Variational Autoencoder (BIR-VAE) provides a meaningful latent representation with an
information resolution that can be specified directly in bits by the system designer. The
rate constraint can be used to prevent overtraining, and the method naturally
facilitates quantisation of the latent variables at the set rate. Our experiments confirm
that the BIR-VAE has a meaningful latent representation and that its performance is at
least as good as state-of-the-art competing algorithms, but with lower computational
complexity. 
\end{abstract}

\section{Introduction}
\label{s:intro}
Generative modelling is an area of machine learning that focuses on discovering the distribution of a data-set. Latent variable models assume there is some collection of underlying information that can characterise the data efficiently. For example, hair colour and facial expression might be a subset of features that describe images of faces. Good representations have numerous applications in machine learning. The effectiveness of machine learning techniques depends on the quality of the data being used as input. Consequently, feature construction/extraction is an important pre-processing step in many machine learning applications \cite{bengio2013representation}. If the features learned by these generative latent feature models represent the essential components of the input data-set, then it may be possible to use them in place of the original data as the input to another machine learning model, such as a classifier. This paper aims to produce a generative model with features that are a meaningful representation of the data.

Variational Autoencoders \cite{kingma2013auto,rezende2014stochastic} (VAEs) and Generative Adversarial Networks \cite{goodfellow2014generative} (GANs) are common latent feature models. However, the latent features produced by the GAN and VAE often are not a good summary of the input. From a representation learning \cite{bengio2013representation} standpoint, these models leave much to be desired. 

A GAN consists of two components, a generator and a discriminator; both are implemented with neural networks. The generator attempts to create fake data that is indistinguishable from real data, and the discriminator attempts to distinguish between the real and fake data, creating a game between the two networks. The generator input is noise $z \sim p(z)$  of a predefined distribution. GANs have been used in a wide variety of tasks, including image-to-image translation \cite{zhu2017unpaired,isola2017image} and image super-resolution \cite{ledig2017photo}. An effort has been made to make the representation of $z$ meaningful \cite{chen2016infogan}.

It has been shown that the GAN objective function is equivalent to minimising the variational lower bound \cite{barber2003algorithm} on the mutual information between the discriminator's input, $x_{Dis}$ and the corresponding labels, $y$ (whether the data is real or fake) \cite{li,huszar_2016}. If $I(x_{Dis}, y) = 0$, then $x_{Dis}$ carries no information about whether the samples are real or fake. However, minimising a lower bound on $I(x_{Dis}, y)$ does not guarantee the quantity will be equal to 0.  This is likely a cause of the instability of the GAN paradigm \cite{li,huszar_2016}.

The Variational Autoencoder is a method for learning generative latent variable models that avoids the problems present with the GAN. The VAE model is defined as $p_\theta(x) = \int_z p(z) \cdot p_\theta(x|z)\ dz$ where $p_\theta(x|z)$ is a distribution implemented using a neural network with parameter $\theta$, and the latent features, $z$, are assumed to be distributed according to $p(z)$, which is pre-defined. Maximising the likelihood of the data given the model is a natural way to train the parameters. However, because of the integral over $z$, the likelihood is typically intractable in a practical implementation. Instead, a lower bound to the likelihood is maximised, called the evidence lower bound, or ELBO. Optimisation of the ELBO induces another distribution $q_\phi(z|x)$, often called the "encoder", which is also implemented with a neural network. Maximising the ELBO corresponds to optimising the likelihood of the data under the model and minimising the Kullback$-$Leibler divergence between $q_\phi(z|x)$ and $p(z)$, where $p(z)$ is still the assumed distribution of latent features (a unit Gaussian is often used \cite{kingma2013auto}). VAEs have been successfully applied to a variety of different problem domains, such as learning to generate handwritten digits \cite{kingma2013auto}, faces \cite{kingma2013auto,kulkarni2015deep} and CIFAR images \cite{gregor2015draw}.

Variational Autoencoders (VAEs) have been criticised because of their inability to learn latent features that are a meaningful representation of the data \cite{zhao2018infovae, phuong2018the, higgins2016beta, chen2016variational}. However, the original formulation of the VAE was to learn a generative model, not to produce latent features that represent the salient information of the data. Recent work into improving the representation learning capabilities of the VAE have adjusted its objective function to reward models with meaningful features \cite{zhao2018infovae, phuong2018the, higgins2016beta}. This paper proposes that it is not enough to modify the VAE architecture, because the original VAE formulation is not concerned with meaningful latent representations. Instead, a new approach is needed.

We propose the Bounded Information Rate Variational Autoencoder (BIR-VAE). The BIR-VAE maximises the likelihood of the data subject to a bound on the information rate that can be conveyed by the latent variables from encoder to decoder. The bound is straightforward to implement by forcing the conditional distribution of the encoder output given the input, $q_\phi(z|x)$, to have a Gaussian distribution with fixed, pre-determined standard deviation. In most scenarios, the bound will be reached, and this implies that the BIR-VAE approach subsumes the objective of mutual information maximisation between the input $x$ and the latent variables $z$ subject to the rate constraint.

The remainder of this paper first surveys recent works which build on the VAE to develop representation learning models, discussing the problems with each. Next, the BIR-VAE is derived; this is done by identifying the criterion that the model should satisfy and subsequently, converting these into a function that can be optimised. Lastly, the BIR-VAE is evaluated experimentally.

\section{Background}
\label{s:background}

This section describes recent work towards creating meaningful latent features in VAEs; thus both motivating our work and providing a context for it. We first discuss in section \ref{s:VAE} the basic VAE \cite{kingma2013auto,rezende2014stochastic}, and then in section \ref{s:mutual} some variants of this method that specifically aim to make the latent features more meaningful. This is generally done by considering the mutual information between the input and the latent variables. The variants include InfoVAE \cite{zhao2018infovae}, the method of the "Fixing a Broken ELBO" paper \cite{alemifixing}, the Mutual Autoencoder \cite{phuong2018the} and the Adversarial Autoencoder \cite{makhzani2015adversarial}.

As is common in work on VAEs, we abuse the formal notation of probability theory. We do not distinguish between random variables and their realisations, assuming that this is clear from the context. We also use the common convention that the argument of a  density labels the density when it is not ambiguous. As an illustration, using both these conventions we can state that $p(z)$ and $p(x)$ describe the densities of the random variables $x$ and $z$. Random variables are real-valued except where stated otherwise.

\subsection{The Variational Autoencoder}
\label{s:VAE}
Variational Autoencoders \cite{kingma2013auto,rezende2014stochastic} (VAEs) are a type of generative latent feature model. That is, they learn a relationship between a set of latent features $z$, and the data $x$. The VAE model is written as $p_\theta(x) = \int_z p(z) \cdot p_\theta(x|z)\ dz$, where $p_\theta(x|z)$ is given by a neural network with parameter $\theta$ and the distribution $p(z)$ is assumed to be simple, e.g., a unit Gaussian \cite{rezende2014stochastic}. Maximising the likelihood of the data under the model is a natural way to train the parameters. However, this is often intractable because of the integral over $z$. Instead, VAEs maximise a lower-bound on the likelihood called the evidence lower bound, or ELBO. The ELBO induces another probability distribution, $q_\phi(z|x)$, which is represented by a neural network with parameters $\phi$. The objective function is
\begin{equation}
\label{equ:elbo}
O_{ELBO} =-D_{KL}[q_\phi(z|x) || p(z)] + \EX_{q_\phi(z|x)}[\log p_\theta(x|z)],
\end{equation}
where $D_{KL}$ is the Kulback-Leibler divergence. In the context of Autoencoders, $p_\theta(x|z)$ can be interpreted as the decoder and $q_\phi(z|x)$ as the encoder. 

The VAE's similarity to an Autoencoder is deceptive. On the surface, one distribution encodes data points into a vectors of latent variables, and another distribution subsequently decodes the latent vectors back into data points. Optimising the ELBO then, in part, maximises the likelihood of the data under the model $p_\theta(x|z)$, and intuitively, $z$ is expected to represent the salient information in the data. However, the notion of an encoder was not present in the original objective, which was to maximise the likelihood of $p_\theta(x)$. Let us write the ELBO in its most basic form, as the sum of a likelihood and a Kullback-Leibler divergence:
\begin{equation}
\label{equ:expanded-elbo}
\begin{split}
O_{ELBO} =& \log p_\theta(x) - D_{KL}[q_\phi(z|x) || p_\theta(z|x)] \\
=& \EX_{z \sim q_\phi(z|x)}[\log p_\theta(x|z)] - D_{KL}[q_\phi(z|x)||p(z)].
\end{split}
\end{equation}
\eqref{equ:expanded-elbo} shows that when the ELBO is maximal, it is equal to the likelihood (assuming the network $q_\phi$ is of sufficient complexity). Consequently, both $D_{KL}[q_\phi(z|x)||p_\phi(z|x)]$ and $D_{KL}[q_\phi(z|x)||p(z)]$ must be 0, which can only occur when $z$ is independent from $x$. If for a given model the ELBO cannot become maximal, then the KL divergence between $q_\phi(z|x)$ and $p(z)$ must be non-zero. Therefore $z$ and $x$ are dependent. Consequently, when $z$ does carry information about $x$ it is because the decoder does not have sufficient complexity to model the data distribution as a function of $p(z)$. This phenomenon was identified when using an LSTM decoder  \cite{bowman2015generating}, and recent works introduce it as the Information Preference Property \cite{chen2016variational,zhao2018infovae,alemifixing,phuong2018the}. In the context of representation learning, the Information Preference Property is problematic. However, the above argument shows that learning salient features of the data was never the purpose of maximising the ELBO.

The second issue is called the Exploding Latent Space problem \cite{zhao2018infovae}, which occurs when the model is sufficiently restrictive, and a larger ELBO can be achieved by maximising the likelihood regardless of the KL divergence term. Optimising $\EX_{q_\phi(z|x)}[\log p_\theta(x|z)]$ maximises the likelihood of observing the data given its corresponding latent variables. Consequently, for a data-set, $\{x_1, ..., x_n\}$, minimising the probability of sampling any $x_j \neq x_i$ but $x_i$ from $p_\theta(x|z)$, where $z \sim q_\phi(z|x_i)$ is a way to increase the likelihood. Therefore, if the distributions $q_\phi(z|x_i)$ have disjoint supports, the decoder can be selected as to map the support of $q_\phi(z|x_i)$ to a distribution centred on $x_i$. This observation shows that maximising the likelihood drives the distributions $q_\phi(z|x_i)$ apart. The KL divergence term regularises this behaviour by pushing the distributions $q_\phi(z|x_i)$ together (towards $p(z)$); however, it is not always successful \cite{zhao2018infovae}. 

The original VAE objective was to learn a generative model of the form $p_\theta(x) = \int_z p(z) \cdot p_\theta(x|z)\ dz$; however, the Exploding Latent Space problem causes the distributions $q_\phi(z|x)$ to diverge rather than converge on $p(z)$. Consequently, the latent variables are not distributed according to $p(z)$, meaning the generative model $p_\theta(x) = \int_z p(z) \cdot p_\theta(x|z)\ dz$ will not produce convincing samples.

This section has shown that a high likelihood (or ELBO) is not indicative of latent features that represent the salient information of the data. Moreover, maximising the ELBO is not suppose to learn a latent representation that is meaningful, because the encoder is a construct of the ELBO and not the original problem formulation. When the model has learned latent features that have captured the data, it is because the decoder is sufficiently restrictive. On the other hand, a high likelihood (and ELBO) can also occur when a poor generative model has been learned. It is worth noting that the quality of $z$ as a representation of $x$ is controlled by the information between $x$ and $z$, which is determined by the joint distribution $q_\phi(z,x)$, something that is not directly affected by maximising the likelihood (and ELBO) \cite{phuong2018the}.

\subsection{Representation Learning based on Mutual Information}
\label{s:mutual}
Mutual information maximisation is an increasingly common method for representation learning, which has been recently applied to both VAEs \cite{zhao2018infovae, phuong2018the, higgins2016beta} and GANs \cite{chen2016infogan}.  In general, mutual information is a measure that has a wide range of applications in neural networks. 

In 1988, Linsker introduced InfoMax \cite{linsker1988self} as a paradigm for optimising Neural Networks. An InfoMax algorithm optimises a function as to maximise the mutual information between the input and output under specified constraints. Particularly well known is the Bell and Sejnovski algorithm \cite{bell1995information} that uses InfoMax to perform Independent Component Analysis.

Recently, mutual information has been used to study the dynamics of learning in deep neural networks \cite{shwartz2017opening, saxe2018information}. The view is that in supervised learning each successive network layer attempts to reduce information about the input while retaining as much information about the desired output as possible. Therefore, the learning network is seen as implementing an approximation to the information bottleneck principle \cite{tishby2000information}. The information bottleneck principle simultaneously minimises the mutual information between the input and the current layer and maximises of the mutual information between the current network layer and the desired output, subject to a relative weighting.  

In the context of VAEs, mutual information is used to ensure that the latent variables $z$ provide useful information about the input $x$. In this subsection, we discuss some approaches to this paradigm in more detail, thus providing a context and a motivation for the BIR-VAE that we introduce in section \ref{s:BLIRVAE}.

\subsubsection{Info Variational Autoencoders}
\label{s:infoVAE}
The family of InfoVAE models \cite{zhao2018infovae} was proposed for solving both the Information Preference Property and the Exploding Latent Space problem that were discussed in section \ref{s:VAE}. Rearranging the ELBO objective function \eqref{equ:elbo} gives the base formula that is modified to find the InfoVAE objective:
\begin{equation}
\label{equ:rearanged-elbo}
\begin{split}
O_{ELBO} &= -D_{KL}[q_\phi(z)||p(z)] \\&+ \EX_{p(z)}[D_{KL}[q_\phi(x|z) || p_\theta(x|z)]].
\end{split}
\end{equation}
The InfoVAE objective function is constructed by adding a scaling term, $\lambda$ to the divergence between $q_\phi(z)$ and $p(z)$ in \eqref{equ:rearanged-elbo}, and adding the mutual information between $x$ and $z$ to the equation with regularisation parameter $\alpha$:
\begin{align*}
O_{InfoVAE} &= - \lambda D_{KL}[q_\phi(z) || p(z)] 
\\ &+ \EX_{q_\phi(z|x)}[\log D_{KL}[q_\phi(x|z) || p_\theta(x|z)]] 
\\ &+ \alpha I_{q}(x;z) \numberthis \label{equ:modified-elbo} \\
&= - \EX_{p_D(x)} \EX_{q_\phi(z|x)}[\log p_\theta(x|z)] 
\\ &- (1 - \alpha)\EX_{p_D(x)} D_{KL}[q_\phi(z|x) || p(z)] 
\\ &- (\alpha + \lambda - 1)D_{KL}[q_\phi(z) || p(z)] \numberthis \label{equ:infovae-pre}
\end{align*}
where $p_{D}(x)$ is the data distribution. \eqref{equ:infovae-pre} gives the InfoVAE's objective. It cannot be optimised directly because of the KL divergence term between $q_\phi(z)$ and $p(z)$. It is proven \cite{zhao2018infovae} that if $\alpha < 1$ and $\lambda > 0$, then $D_{KL}[q_\phi(z) || p(z)]$ can be replaced with any strict divergence between $q_\phi(z)$ and $p(z)$. Consequently, it is possible to use the Maximum Mean Discrepancy \cite{gretton2012kernel} as the divergence; this model is named the MMD-VAE. 

When $\alpha \not= 1$, $D_{KL}[q_\phi(z|x) || p(z)]$ and $D_{KL}[q_\phi(z) || p(z)]$ are being simultaneously minimised. This pair is optimal only if $z$ is independent from $x$. However, the original objective \eqref{equ:modified-elbo} was formulated to maximise the mutual information between $z$ and $x$. Consequently, the InfoVAE objective is penalising the model when $I_{q_\phi}(x;z) > 0$, while maximising $I_{q_\phi}(x;z)$.

\subsubsection{Adversarial Autoencoder}
The Adversarial Autoencoder (AAE) \cite{makhzani2015adversarial}, is structured like a VAE, except instead of minimising the KL divergence between $q_\phi(z|x)$ and $p(z)$, it uses an adversarial training technique to drive the distribution $q_\phi(z)$ towards $p(z)$, where $p(z)$ is a predetermined distribution, same as for a VAE. Samples taken from $p(z)$ are considered the real data and latent variable vectors produced by the encoder network are considered fakes, an additional neural network is constructed which is used to discriminate between samples from $p(z)$ and the latent variable vectors. The encoder network is penalised if the discriminator can tell that the vector of latent features did not come from $p(z)$, so it is driven to produce latent codes that are distributed according to $p(z)$. 

The Adversarial Autoencoder is also part of the family of InfoVAEs. This can be seen by taking \eqref{equ:infovae-pre}, letting $\alpha = \lambda = 1$ and using the Jensen-Shannon divergence between $q_\phi(z)$ and $p(z)$ \cite{zhao2018infovae}. Consequently, AAEs do not suffer from the same problems that a standard VAE does \cite{zhao2018infovae}. However, training GANs can be unstable, and with new methods to improve stability, it can be slow \cite{arjovsky2017wasserstein,gulrajani2017improved}. Consequently, other methods are preferable to the AAE \cite{zhao2018infovae}.

\subsubsection{Fixing a Broken ELBO}
As discussed previously, maximising the ELBO is not sufficient for representation learning as it gives no guarantees that $z$ will contain any information about $x$. Moreover, maximising the ELBO encourages $z$ to be independent of $x$. \cite{alemifixing} makes use of variational upper and lower bounds on the mutual information to prevent this behaviour.

Consider again the data to be $x$. An "encoding" distribution $q_\phi(z|x)$ takes data vectors and produces a distribution over latent representations. The encoder induces two distributions of interest, $q_\phi(z)$ and $q_\phi(x|z)$, both of which cannot be computed. Consequently, $p_\theta(x|z )$ and $m_\omega(z)$ are introduced, which are approximations of $q_\phi(x|z)$ and $q_\phi(z)$ respectively. 

The encoding channel represented by the distribution $q_\phi(z|x)$ has a maximum amount of information that can be transferred through it, denoted $R$. Consequently, $I(x;z) \leq R$ because $z$ cannot contain more information about $x$ than can be put through the encoding channel. The mutual information is bounded from below by the entropy of $x$ minus reconstruction likelihood.
\begin{equation}
\label{equ:bounded-mutual-info}
\begin{split}
H(x) - \EX_{p_D(x)} \EX_{q_\phi(z|x)}[\log p_\theta(x|z)] \leq I_{q_\phi}(x;z) \\ \leq \EX_{p_D(x)}[D_{KL}[q_\phi(z|x) || m_\omega(z)]].
\end{split}
\end{equation}
(Following \cite{alemifixing} we implicitly assume discrete variables.) \eqref{equ:bounded-mutual-info} demonstrates the bounds on the mutual information. To train this model either the upper or lower bound is regularised to stay at a predetermined value while the other is optimised \cite{alemifixing}. However, this solution does require computing both the distortion and rate, increasing the complexity of the optimisation problem. Moreover, regularising the model to have a preferred rate does not guarantee that this condition will be met.

\subsubsection{Mutual Autoencoder}
The Mutual Autoencoder \cite{phuong2018the} is another approach which uses the mutual information to ensure a meaningful latent representation is learned. It regularises $I_\theta(x, z)$ to keep it at a pre-specified value. However, given the difficulty of computing the mutual information, an approximation (lower bound) is used instead. The approximation of $I_\theta(X, Z)$ is given by $I_\theta(X, Z) \geq \hat{I}_\theta(X, Z) = H(z) + \EX[\log r(z|x)]$, where r(z|x) is any conditional distribution; this is called the Variational Infomax bound \cite{barber2003algorithm}. 

The distribution $r(z|x)$ is an auxiliary model that must also be trained, increasing the complexity of training and the number of parameters. Another concern is that if $r(z|x)$ is $p_\theta(z|x)$ then $I_\theta(X, Z) = \hat{I}_\theta(X, Z)$, but if $r(z|x)$ is not a good approximation of $p_\theta(z|x)$ then $\hat{I}_\theta(X, Z)$ is not a good measure of $I_\theta(X, Z)$. In other words, there are no guarantees on how tight the bound on the mutual information is. The Mutual Autoencoder has promising experimental results; however, the authors report that the Mutual Autoencoder is slow to train because it requires computing the additional mutual information term.
\section{Bounded Information Rate VAE}
\label{s:BLIRVAE}
This section first derives the Bounded Information Rate VAE (BIR-VAE) by describing the induced information rate bound on the encoding channel, and then deriving an objective function that can be optimised. A theoretical comparison between the BIR-VAE and other recent works is also given, describing the contribution of the BIR-VAE in relation to the existing work.

\subsection{Theory}
\label{s:BLIRVAE:theory}
The fundamental principle of BIR-VAE is to define an objective function that maximises the likelihood that the input is observed at its output (similar to basic VAE \cite{kingma2013auto, rezende2014stochastic}), subject to a constraint on the information rate flowing through the latent layer. The objective and the constraint naturally lead to a meaningful representation with any desired resolution of information about the input. Importantly, as we will show below, this simple paradigm enforces the mutual information between the latent variables $z$ and the input $x$, without requiring the computation of the mutual information.

A distinction is made between the output of the encoder prior to the channel, $y$, and the latent variables, $z$, that form the output of the channel. Hence $y$ corresponds to the mean of the latent variable distribution in a conventional VAE. The desired distribution of the latent variables $q_\phi(z)$ is defined as iid Gaussian with unit variance in each dimension. If $x$ is the random input vector then the deterministic network $\mu_\phi(\cdot)$ transforms $x$ into $y$, the mean of the distribution $q_\phi(z|x)$. Noise is added to $y$, to throttle the information throughout the latent variables; this gives, $z$: $z=y+\epsilon$, where $\epsilon$ is iid Gaussian noise with variance $\sigma_\epsilon^2<1$ for each dimension.  The variance $\sigma_\epsilon^2$ is set by the system designer and determines the information rate. Note that this differs from the conventional VAE, where the variance $\sigma^2_\epsilon$ is learned, and the information rate is unknown. The information rate of BIR-VAE across the channel is now bound to 
\cite{cover2012elements}  
\begin{equation}
\label{q:BIRVAErate}
I = \frac{d}{2}\,\log_2(\frac{1}{\sigma_\epsilon^2}),
\end{equation}
where $d$ is the dimensionality of the latent layer.

To show that BIR-VAE subsumes maximising the mutual information between the latent variables $z$ and the input $x$ we consider an objective function that maximises
\begin{enumerate}
\item the likelihood of the input to be seen at the output (similar to basic VAE \cite{kingma2013auto} \cite{rezende2014stochastic}), and
\item the mutual information between the latent variables and the input (similar to InfoVAE \cite{zhao2018infovae}, and to the Mutual Autoencoder \cite{phuong2018the}),
\end{enumerate}
subject to a constraint on the information rate flowing through the latent layer. The constrained information rate is induced by placing two restrictions on the latent configuration. Firstly, the distribution $q_\phi(z)$ is defined to be $N(0, I)$. Secondly, $q_\phi(z|x)$ is defined as a Gaussian distribution with arbitrary mean and a variance of $\sigma_\epsilon^2$ in each dimension.

Let $p_D(x)$ be the data distribution over x,  and let $q_\phi(z|x)$ and  $p_\theta(x|z)$ be the encoder and decoder respectively. Furthermore, let $I_{q_\phi}(x;z)$ be the  mutual information between x and z under the joint distribution $q_\phi(x, z)$. Then we have
\begin{equation}
\label{equ:blir-vae-objective-contrained}
\begin{aligned}
&\underset{\phi, \theta}{\text{max}}&&\EX_{p_D(x)} \EX_{q_\phi(z|x)}[\log p_\theta(x|z)] +
\omega I_{q_\phi}(x;z)\\
&\text{subject to} &&q_\phi(z) = N(0, I),\\
&&& \EX_{q_\phi(z|x)}[ (x - \EX_{q_\phi(z|x)}[z])^2 ] =   \sigma^2_\epsilon I,
    \end{aligned}
\end{equation}
where $\sigma^2_\epsilon$ is a variance set by the system designer that determines the rate constraint, and $\omega$ is a weighting. It is possible to satisfy the second constraint, $\EX_{q_\phi(z|x)}[ (x - \EX_{q_\phi(z|x)}[z])^2 ] =   \sigma^2_\epsilon I$ by fixing the amount of noise introduced in the latent layer. 

The second constraint, $q_\phi(z) = N(0, I) = p_\theta(z)$, is satisfied when the Maximum Mean Discrepancy \cite{gretton2012kernel} between $q_\phi(z)$ and $p(z)$ is 0. We can form a Lagrangian an write the BIR-VAE objective \eqref{equ:blir-vae-objective-contrained} as 
\begin{equation}
\label{equ:blir-vae-objective}
\begin{aligned}
&\underset{\phi, \theta}{\text{max}}&&\begin{split}\EX_{p_D(x)} \EX_{q_\phi(z|x)}[\log
  p_\theta(x|z)] + \omega I_{q_\phi}(x;z)\\ - \lambda MMD[q_\phi(z) || N(0, I)]\end{split}\\
&\text{subject to}&& \EX_{q_\phi(z|x)}[(z - \EX_{q_\phi(z|x)}[z])^2 ] =   \sigma^2_\epsilon I.
 \end{aligned}
\end{equation}
To optimise \eqref{equ:blir-vae-objective} the term
  $I_{q_\phi}(x;z)$ must be made tractable. A convenient form for the mutual information is
\begin{equation}
    \label{equ:var-info-bound}
    I_{q_\phi}(x;z) = h_{q_\phi(z)}(z) + \EX_{p_D(x)}[ h_{q_\phi(z|x)}(z)],
\end{equation}
where $h$ denotes differential entropy. We note that if the constraint
 $q_\phi(z) = N(0, I)$ is satisfied, then the differential entropy  $h_{q_\phi}(z)$ is fixed. Hence the
differential entropy $h_{q_\phi(z)}(z)$ can be omitted from the  optimisation problem. The differential entropy $h_{q_\phi(z|x)}(z) $ is the entropy of the latent  variable $z$ for a given input $x$. This is  also fixed as this
conditioned variable has a Gaussian  distribution with variance $\sigma_\epsilon^2$. Consequently, the second 
term of the mutual information can also  be omitted.

We have now show that it is possible to write the BIR-VAE objective as
\begin{equation}
\label{equ:blir-vae-objective-optimisable}
\begin{aligned}
&\underset{\phi, \theta}{\text{max}}&&\begin{split}&\EX_{p_D(x)} \EX_{q_\phi(z|x)}[\log
  p_\theta(x|z)] \\&- \lambda MMD[q_\phi(z) || N(0, I)] \end{split}\\
&\text{subject to}&& \EX_{q_\phi(z|x)}[ (z - \EX_{q_\phi(z|x)}[z])^2 ] =   \sigma^2_\epsilon I.
    \end{aligned}
\end{equation}

\begin{figure}
\includegraphics[width=0.5\textwidth]{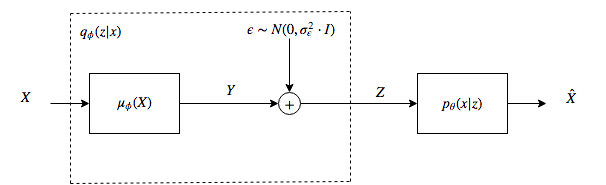}
\caption{Architecture of the Bounded Information Rate VAE. The encoder network, $\mu_\phi(X) = y$, outputs the mean of the distribution $q_\phi(z|x)$, then noise $\epsilon \sim N(0, \sigma_\epsilon^2)$ is added to $y$ to get the latent variables $z$.}
\label{fig:blir-vae-arch}
\end{figure}

Figure \ref{fig:blir-vae-arch} shows the structure of the model, which is similar to the
implementation of a VAE except that the variance of $z$ given $x$ is not
computed by the BIR-VAE encoder, because it is a constant.

The BIR-VAE decoder
($P_\theta(x|z)$) outputs a distribution, however, if the output distribution is assumed
to be an isotropic Gaussian (i.e. $N(0, \sigma^2 \cdot I)$), then the decoder produces as
output simply the mean of an isotropic Gaussian with $\sigma^2 = 1$
\cite{kingma2013auto}. This reduces the log likelihood to the negative mean square error:
\begin{equation}
    \begin{aligned}
        &\log \big[\mathrm{det}(2 \pi \Sigma) \cdot e^{-\frac{1}{2} (x - \mu)^T \Sigma^{-1}(x - \mu)} \big]\\
        &=\log \big[\mathrm{det}(2 \pi \Sigma)\big] + -\frac{1}{2} (x - \mu)^2,
    \end{aligned}
    \label{equ:rearanged-log-pdf}
\end{equation}
where $\Sigma, \mu$ are the covariance matrix and mean, respectively.  In a practical
implementation we maximise \eqref{equ:rearanged-log-pdf} over a batch and $\log[\mathrm{det}(2 \pi \Sigma)]$ is ignored (because it is constant). The parameter $\lambda$ can be simply set to a value that ensures the MMD between
$q_\phi(z)$ and $N(0, I)$ is on a similar scale to the likelihood error. 

Algorithm \ref{a:blirvae} summarises the method. In the Algorithm capital letters denote
data sets equivalent to a minibatch and $MSE$ denotes mean squared error. 

\begin{algorithm}[!h]
\SetAlgoLined
\caption{The BIR-VAE algorithm. \label{a:blirvae}}
\scriptsize{
\KwData{Input signal $\{x_i\}$}
\KwResult{Optimised parameters $\theta^*, \phi^*$ for encoder and decoder}
set variance $\sigma_\epsilon^2$ of distribution $q_\phi(z|x)$ \;
set weight $\lambda$ \;
initialise parameters $\theta, \phi$\;
\For{ each epoch $l\in\mathcal{L}$ }{
  \For{ each minibatch $n\in\mathcal{N}$ }{
    $X_l \leftarrow $ current minibatch \;
    $Y \leftarrow \mu_\phi(X_l)$ \hspace{1em} \% encoder\;
    $Z \leftarrow Y + \epsilon, \,\,\epsilon \sim N(0, \sigma_\epsilon^2)$  \hspace{1em} \% channel\;
    $\hat{X} \leftarrow p_\theta(\cdot|Z)$  \hspace{1em} \% decoder\;
    $L \leftarrow MSE(\hat{X}, X_l) + \lambda MMD[q_\phi(Z) || N(0, I)]$ \;
    $(\theta, \phi) \leftarrow$ +Adam update of $\theta, \phi$ to minimise $L$\;
  }
}
}
\end{algorithm}

\subsection{Discussion}
The approach described in \ref{s:BLIRVAE:theory} can be seen as a more tractable method to reach the goals of the Mutual Autoencoder \cite{phuong2018the}. Instead of attempting to fix the mutual information of $z$ and $x$ through a penalty term, the BIR-VAE physically restricts the information rate of the encoding channel. Where the Mutual Autoencoder requires computing the mutual information of $z$ and $x$ (so that it can be regularised), the BIR-VAE avoids this by building the information rate restriction into the model. By this same argument, the BIR-VAE also improves upon the solution presented in \cite{alemifixing}.

The BIR-VAE objective function, \eqref{equ:blir-vae-objective-optimisable}, can be seen as a special case of the InfoVAE objective, i.e. for the case $\alpha = 1$. However, it is not possible to use the InfoVAE objective when $\alpha = 1$ without restricting the mutual information between $x$ and $z$ under the encoding distribution. Without any restriction in place, the mutual information can be maximised to infinity by making $q_\phi(z|x)$ a deterministic mapping. The authors of the InfoVAE paper note this, and state that ensuring that the variance of $q_\phi(z|x)$ does not approach 0 is sufficient to prevent this behaviour. They do not discuss how this is achieved. Furthermore, by making this restriction, they are inducing a  maximum information rate on the encoding channel, something that is not identified. 

In contrast to existing VAEs \cite{kingma2013auto,rezende2014stochastic}, the variance of the noise $\epsilon$ is pre-determined. In traditional VAEs, the variance of the noise was allowed to vary across the domain of the latent layer variable. However, by scaling the means (or $y$ values in the case of the BIR-VAE) across their domain, the same result is obtained. Hence, for a sufficiently flexible encoder and decoder, the ability to vary the noise variance across the domain is unlikely to affect performance significantly. Conventional VAEs can create a low SNR across the channel for all latent variables; this is the reason why a VAE can ignore the information arriving through the channel and maximise the likelihood using only the decoder. 

The ability to set the channel rate of the BIR-VAE clarifies a disadvantage of the basic VAE structure. While VAEs attempt to  provide a good likelihood for observations, it has no good reason to provide good performance between observed data other than that it provides a reasonable interpolation across the latent variables. In a BIR-VAE, the quality of this interpolation is dependent on the rate. In contrast, in a conventional VAE, the quality of the interpolation is uncertain. From a generative perspective, it is advantageous to set the rate of the BIR-VAE high. However, a high rate requires a larger database for training. For example, to get texture correct, a very high rate likely is required. Note that this differentiates VAEs from GANs. In GANs the generator performance is judged by a discriminator independently of data points seen. As the discriminator can rely on feature extraction for its judgement (for example for texture), it is less dependent on having seen similar data before. 

Finally we note that the BIR-VAE naturally leads to an encoder-decoder system with a quantised bit stream that can be stored or transmitted. In this case the channel is replaced with a vector quantiser, e.g., \cite{gray2012source} with the noise characteristics that approximate the Gaussian distribution of the additive noise $\epsilon$. As the data has a well-defined Gaussian distribution, a lattice quantiser, e.g., \cite{erez2005lattices} is particularly natural. 

\section{Experiments}
\label{s:experiments}

In this section, the performance of the BIR-VAE algorithm is evaluated on the MNIST \cite{lecun1998mnist} and SVHN \cite{netzer2011reading} datasets. The meaningfulness of the latent variables and the effect of the information rate is investigated. As a reference system, the InfoVAE algorithm \cite{zhao2018infovae} is used. In this section the unit bpi refers to bits/image.

\subsection{Experimental Setup and Reference System}

The implementation of the BIR-VAE algorithm follows Algorithm \ref{a:blirvae}. The Maximum Mean Discrepancy measure uses a Gaussian kernel, 
\begin{equation}
    k(x, x^{'}) = e^{- \frac{||x - x^{'}||^2}{2\sigma_{k}^2}}    ,
\end{equation}
with a variance $\sigma_k = 1$. The information rate of the BIR-VAE is set using \eqref{q:BIRVAErate}. That is, the variance of $q_\phi(z|x)$ is set to $\sigma_\epsilon^2 = 1/(4^{\frac{I}{d}})$, where $d$ is the dimensionality of $z$ and $I$ the information rate. The dimensionality $d$ of the latent variables was varied in the experiments.

As a reference system we used the InfoVAE algorithm \cite{zhao2018infovae} described in section \ref{s:infoVAE}. It was selected as representative of state-of-the-art performance and because code written by the authors is available.\footnote{\url{https://github.com/ShengjiaZhao/InfoVAE/blob/master/mmd_vae_eval.py}} We used parameter settings $\alpha = 0$ and $\lambda = 1000$ as this is what the InfoVAE authors used when training on the MNIST dataset. Out of interest, $\alpha = 0.9$ will also be used, as this setting of $\alpha$ means the model prefers a larger mutual information. The setting of $\lambda = 1000$ was also used for the BIR-VAE model, except in one experiment where it is necessary to increase the regularisation constant to enforce the constraint that $q_\phi(z) = N(0, I)$. The InfoVAE does not explicitly define an upper bound on the mutual information. However, the authors of the InfoVAE note that ensuring $q_\phi(z|x)$ does not have vanishing variance is sufficient to regularise the behaviour of the model \cite{zhao2018infovae}. In the code provided by the authors the standard deviation of the conditional latent variable distribution, $q_\phi(z|x)$, is bounded by $\sigma_\epsilon \geq 0.01$; this regularisation was also used in our implementation of InfoVAE.

From comparison to the BIR-VAE algorithm, we note that the bounding of $\sigma_\epsilon$ in InfoVAE corresponds to an implicit bounding of the information rate.  According to \eqref{q:BIRVAErate}, this maximum rate is approximately $13.3$ bpi with a two-dimensional latent space.  Importantly, if this bound is set without consideration of the database size, over-fitting may result. 

We used the MNIST database of hand-written digits \cite{lecun1998mnist} 
and the Street View House Numbers \cite{netzer2011reading} (SVHN) dataset for our experiments.
The MNIST database has 60,000 training images and 10,000 testing images. The data were
used in their native form of 28$\times$28 images with 32-bit intensities.

The Street View House Numbers \cite{netzer2011reading} (SVHN) dataset is a collection of
house numbers that have been segmented into individual digits. The SVHN database consists of 73,257 training samples and 26,032 testing samples. SVHN is more complex than
MNIST because the images are larger (32$\times$32) and in colour. It is worth noting that
the SVHN images contain more than just the subject (number), they have distracting
components around the edges.

\subsection{Results}
We studied the relation between information rate and reconstructive and generative performance. We also studied the descriptiveness of the latent variables with respect to the input.

\subsubsection{Effect of Information Rate}
\begin{figure}
    \centering
    
    \begin{subfigure}{0.49\columnwidth}
        \includegraphics[width=\textwidth]{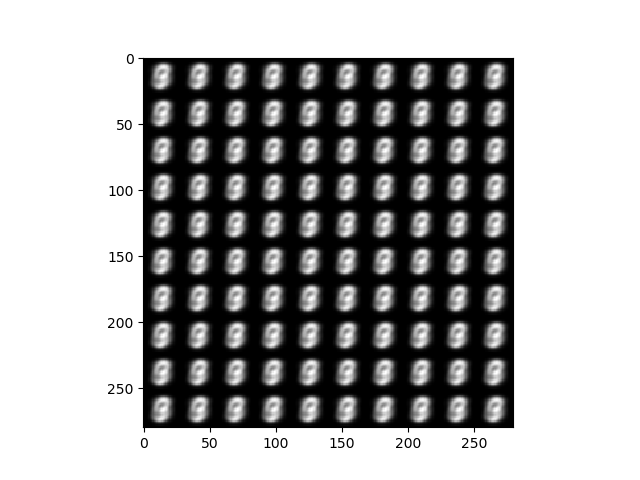}
        \caption{Rate: 0 bpi.}
        \label{fig:blir_vae_2bits_recon}
    \end{subfigure}
    \begin{subfigure}{0.49\columnwidth}
        \includegraphics[width=\textwidth]{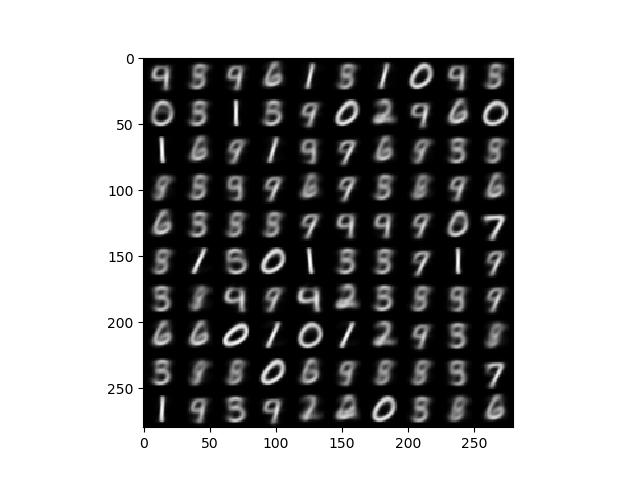}
        \caption{Rate: $\approx$ 3 bpi.}
        \label{fig:blir_vae_6bits_recon}
    \end{subfigure}
    
    \begin{subfigure}{0.49\columnwidth}
        \includegraphics[width=\textwidth]{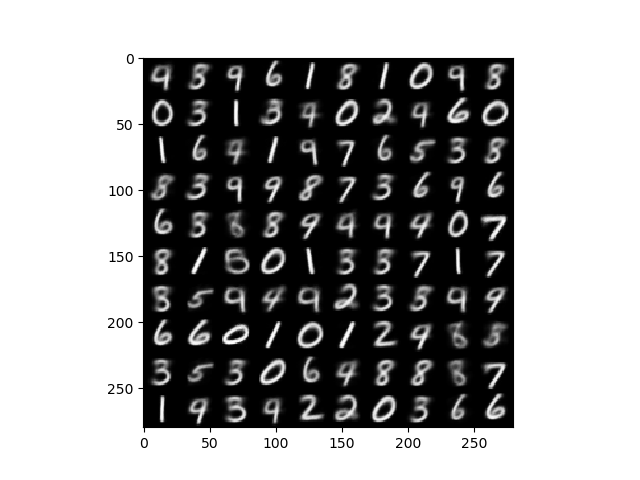}
        \caption{Rate: $\approx$ 7 bpi.}
        \label{fig:blir_vae_14bits_recon}
    \end{subfigure}
       \begin{subfigure}{0.49\columnwidth}
           \includegraphics[width=\textwidth]{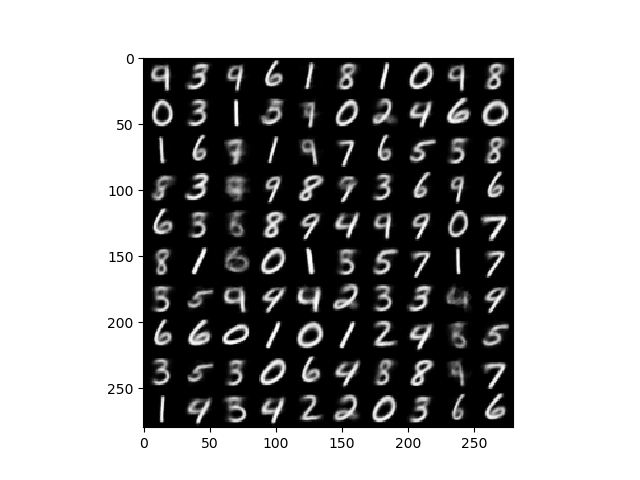}
        \caption{Rate: $\approx$ 13.3 bpi.}
        \label{fig:mnist:blir_vae_26bits_recon}    
    \end{subfigure}

    \caption{Digit reconstructions for the BIR-VAE model with varying information rates.}
    \label{fig:blir_vae_reconstructions}

\end{figure}
\begin{figure}
    \centering
    
    \begin{subfigure}{0.49\columnwidth}
        \includegraphics[width=\textwidth]{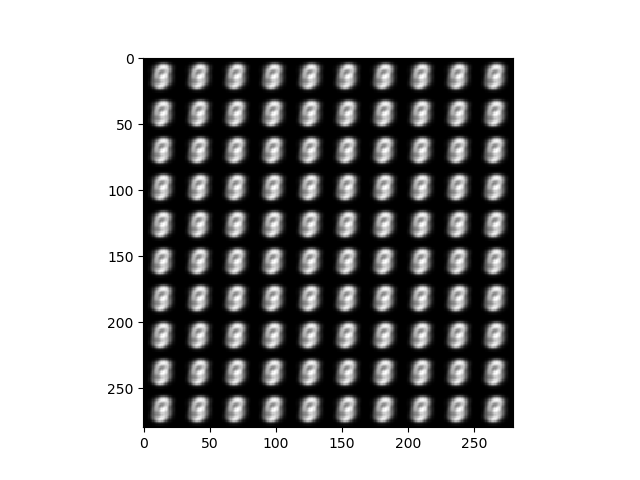}
        \caption{Rate: 0 bpi.}
        \label{fig:blir_vae_2bits_gen}
    \end{subfigure}
    \begin{subfigure}{0.49\columnwidth}
        \includegraphics[width=\textwidth]{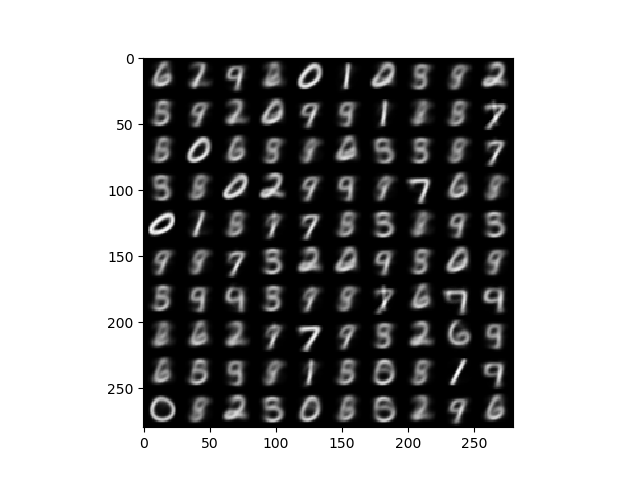}
        \caption{Rate: $\approx$ 3 bpi.}
        \label{fig:blir_vae_6bits_gen}
    \end{subfigure}
    \begin{subfigure}{0.49\columnwidth}
        \includegraphics[width=\textwidth]{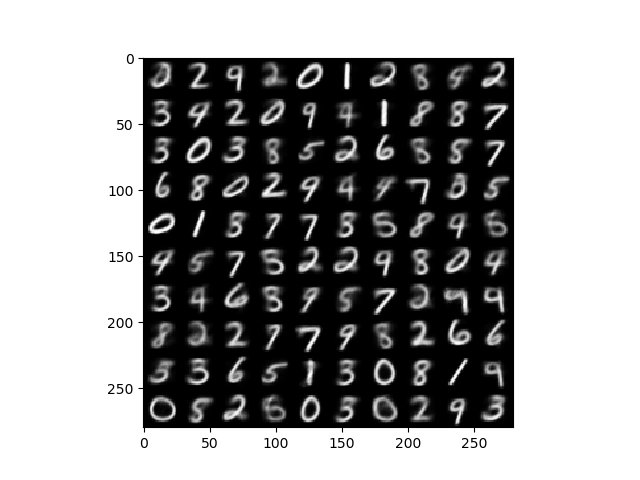}
        \caption{Rate: $\approx$ 7 bpi.}
        \label{fig:blir_vae_14bits_gen}
    \end{subfigure}
    \begin{subfigure}{0.49\columnwidth}
           \includegraphics[width=\textwidth]{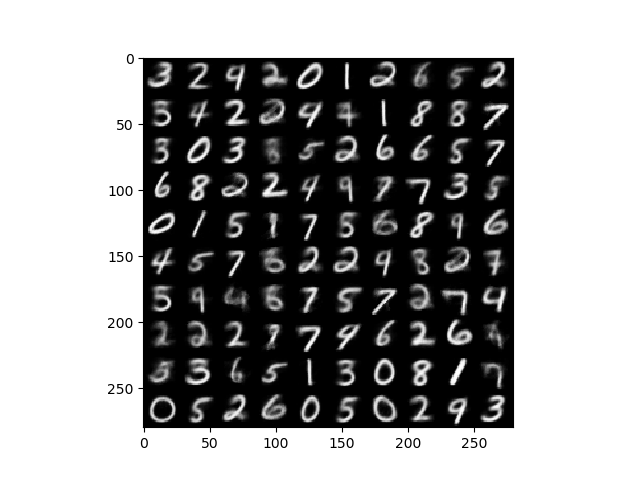}
        \caption{Rate: $\approx$ 13.3 bpi.}
        \label{fig:mnist:blir_vae_26bits_gen}    
    \end{subfigure}

    \caption{Digits generated from BIR-VAEs with varying information rates.}
    \label{fig:blir_vae_generated}
\end{figure}
Figures \ref{fig:blir_vae_reconstructions} and \ref{fig:blir_vae_generated} respectively
show the reconstructions and generations from a BIR-VAE with different information rate
limits on the encoding channel for two latent features ($d=2$). As the information rate
increases, the quality of both the reconstructions and generations increases. It should be noted that when the BIR-VAE's information rate was restricted to $\approx$ 3 bpi,  the hyperparameter $\lambda$ had to be increased to 10,000 to enforce the constraint that $q_\phi(z)$ is a unit Gaussian.

\begin{figure}
    \centering
    \begin{subfigure}{0.49\columnwidth}
        \includegraphics[width=\textwidth]{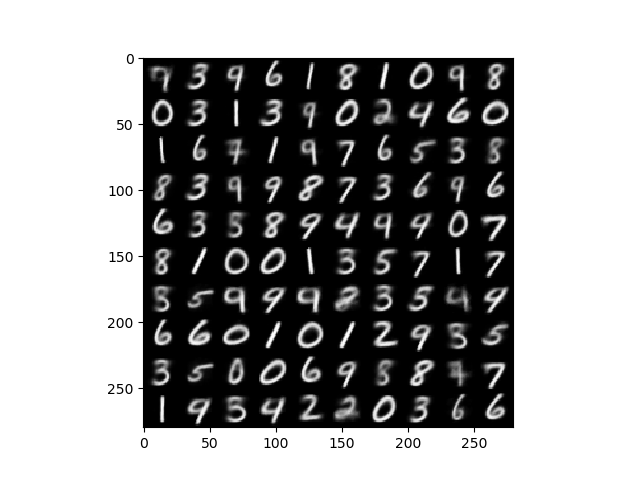}
        \caption{Reconstructed digits.}
        \label{fig:mnist:info_vae_2lf_recon}
    \end{subfigure}
    \begin{subfigure}{0.49\columnwidth}
        \includegraphics[width=\textwidth]{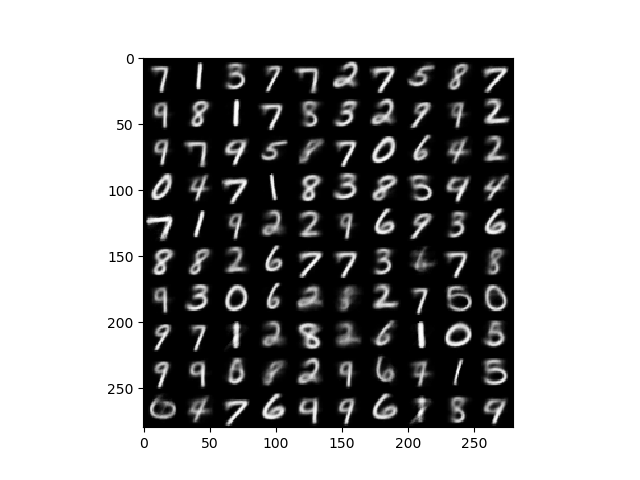}
        \caption{Generated digits.}
        \label{fig:mnist:info_vae_2lf_gen}
    \end{subfigure}
    \caption{Figures generated using an InfoVAE model trained on the MNIST dataset.}
\end{figure}

Reconstructions and generations produced by the InfoVAE model are shown in figures
\ref{fig:mnist:info_vae_2lf_recon} and \ref{fig:mnist:info_vae_2lf_gen}, respectively. The
BIR-VAE and InfoVAE are indistinguishable in quality when BIR-VAE has a rate that is
identical to the maximum rate of the InfoVAE (as noted, $\approx$ 13.3 bpi if $d = 2$). 

\begin{table}
    \begin{tabular}{| c | c | c |}
        \hline
        Model & Train MSE & Train MSE \\
        \hline
        BIR-VAE (0 bpi) & 52.73 & 52.89 \\
        BIR-VAE ($\approx$ 3 bpi) & 37.04 & 36.98 \\
        BIR-VAE ($\approx$ 7 bpi) & 27.78 & 29.79 \\
        BIR-VAE ($\approx$ 13.3 bpi) & 26.01 & 31.38 \\
        InfoVAE ($\alpha = 0$) & 26.97 & 29.85 \\
        InfoVAE ($\alpha = 0.9$) & 28.68 & 30.74 \\
        \hline
    \end{tabular}
    \caption{Mean Square Error (MSE) for various models trained on the MNIST dataset.}
    \label{table:mnist-mse}
\end{table}

Table \ref{table:mnist-mse} shows the training and testing MSE for each of the models trained on the MNIST dataset. As expected, when the information rate is increased, the reconstruction MSE decreases. Both the InfoVAE models have similar performance in this situation.
\begin{figure*}
    \centering
    \begin{subfigure}[t]{0.28\textwidth}
        \includegraphics[width=\textwidth]{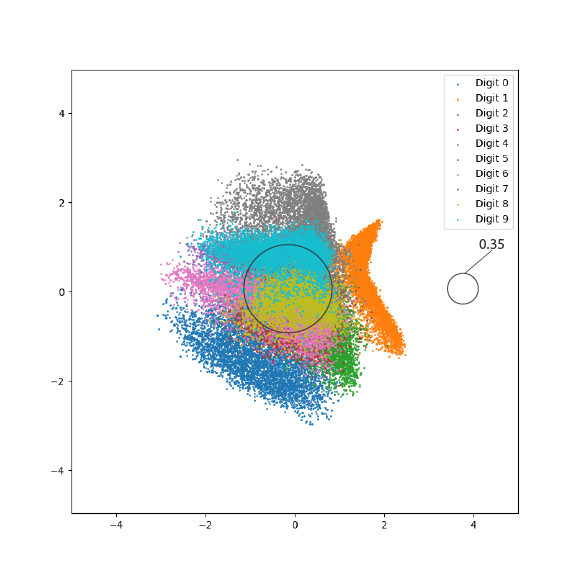}
        \caption{BIR-VAE with Information Rate of $\approx$ 3 bpi. The radius of
          large and small circles represent $\sigma_z$ for $q_\phi(z)$ and
          $\sigma_\epsilon$ for $q_\phi(z|x)$ respectively.}
        \label{fig:blir_vae_6bits_latent_space}
    \end{subfigure}
\hspace{2em}
    \begin{subfigure}[t]{0.28\textwidth}
        \includegraphics[width=\textwidth]{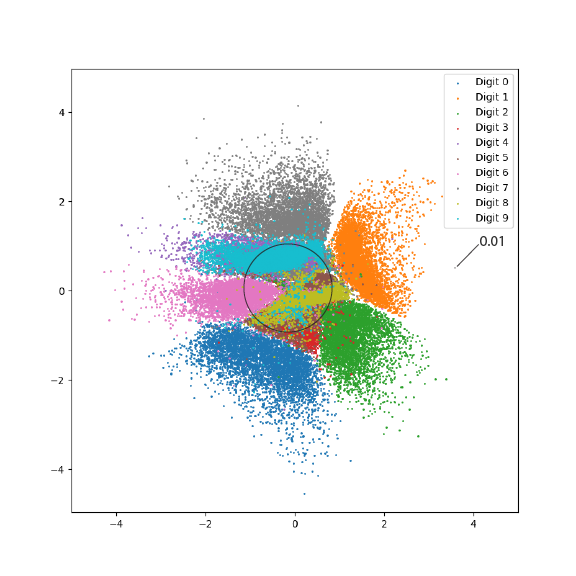}
        \caption{BIR-VAE with Information Rate of $\approx 13.3$ bpi. The radius of
          large and small circles represent $\sigma_z$ for $q_\phi(z)$ and
            $\sigma_\epsilon$ for $q_\phi(z|x)$ respectively. It is difficult to see the circle representing $\sigma_\epsilon$ as it is very small.}
        \label{fig:blir_vae_26bits_latent_space}
    \end{subfigure}
\hspace{2em}
    \begin{subfigure}[t]{0.28\textwidth}
        \includegraphics[width=\textwidth]{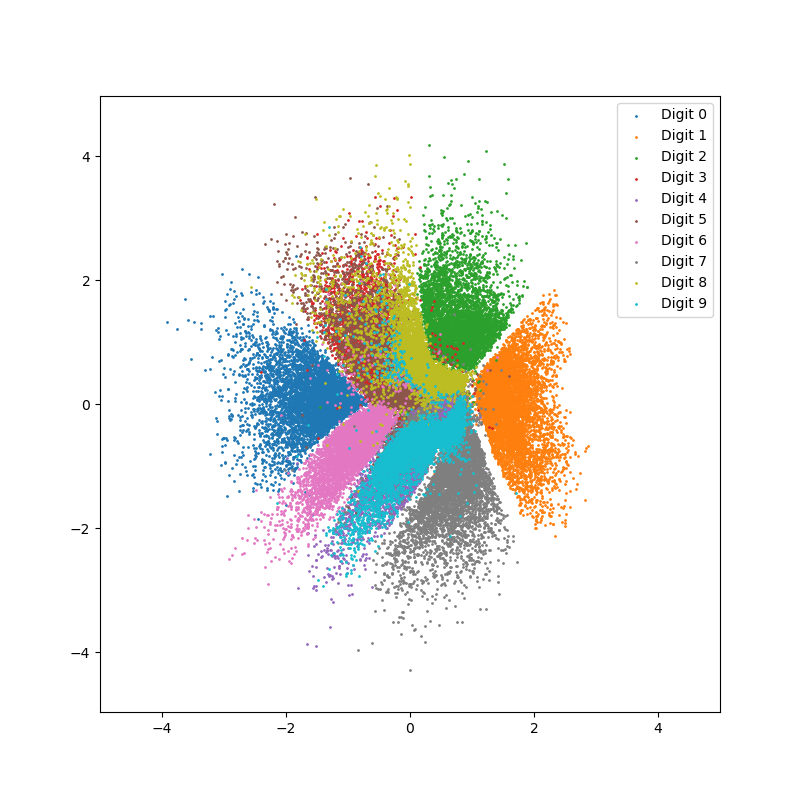}
        \caption{InfoVAE.\\\mbox{}}
        \label{fig:info_vae_2lf_latent_space}
    \end{subfigure}
    \caption{Latent space plots for BIR-VAE and InfoVAE models.}
    \label{fig:mnist:latent-space-plot}
\end{figure*}

As the aim of BIR-VAE is to obtain a meaningful latent representation it is useful to inspect how the information rate affects the organisation of the latent layer. This is particularly straightforward for the case with only two latent dimensions. Figure \ref{fig:mnist:latent-space-plot} shows the latent variables for the InfoVAE and BIR-VAE with $\approx$ 3 and $\approx 13.3$ bits of information per image. The figure shows that the BIR-VAE with $\approx 13.3$ bits of information per image has sharper boundaries between classes than the BIR-VAE with a information rate of $\approx$ 3 bpi. As might be expected, the InfoVAE has a latent representation similar to that of the BIR-VAE with a $\approx 13.3$ bpi information rate. Sharper boundaries between classes mean that the model better understands the differences between the digit classes. 

\subsubsection{Avoiding Overfitting when Data is Limited}
To study overfitting, we used the first 600 elements of the MNIST training data for
training only. Again, we use a two-dimensional latent space.

Table \ref{table:reduced-mnist-mse} shows the MSE for five different models trained on the Reduced MNIST dataset. The table shows how adjusting the information rate of the BIR-VAE can be used to control the overfitting. The discrepancy between the training and test MSE is lowest for the BIR-VAE model with an information rate of $\approx$ 2 bits/image. In contrast, the InfoVAE shows clear signs of overfitting, with a large discrepancy between the performance for the training and testing databases.

\begin{table}
\centering
    \begin{tabular}{| c | c | c |}
        \hline
        Model & Train MSE & Test MSE \\
        \hline
        BIR-VAE($\approx$ 2 bpi) & 35.23 & 42.82 \\
        BIR-VAE ($\approx$ 3 bpi) & 29.38 & 41.46 \\
        BIR-VAE ($\approx$ 5 bpi) & 22.10 & 42.97 \\
        InfoVAE ($\alpha = 0$) & 10.88 & 60.54 \\
        InfoVAE ($\alpha = 0.9$) & 12.08 & 61.68 \\
        \hline
    \end{tabular}
    \caption{Mean Square Error (MSE) for the models trained on the reduced MNIST problem.}
    \label{table:reduced-mnist-mse}
\end{table}

\begin{figure}
    \centering
    \begin{subfigure}{0.49\columnwidth}
        \includegraphics[width=\textwidth]{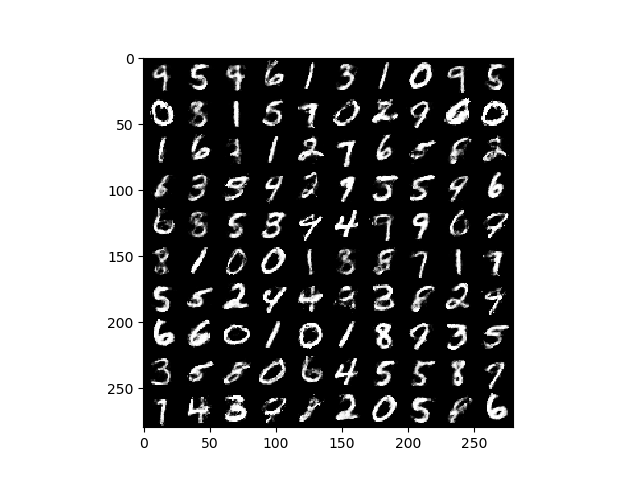}
        \caption{Reconstructed digits.}
        \label{fig:reduced-mnist:info_vae_2lf_recon}
    \end{subfigure}
    \begin{subfigure}{0.49\columnwidth}
        \includegraphics[width=\textwidth]{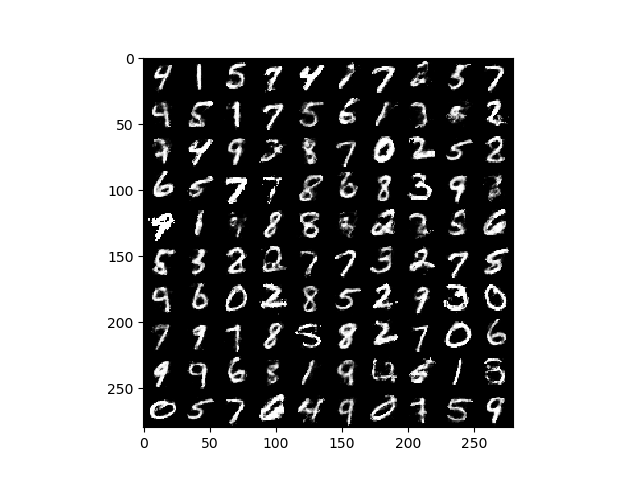}
        \caption{Generated digits.}
        \label{fig:reduced-mnist:info_vae_2lf_gen}
    \end{subfigure}
    \caption{Figures taken from an InfoVAE model trained on a 600 element subset of MNIST.}
\end{figure}

\begin{figure}
    \centering
    \begin{subfigure}{0.49\columnwidth}
        \includegraphics[width=\textwidth]{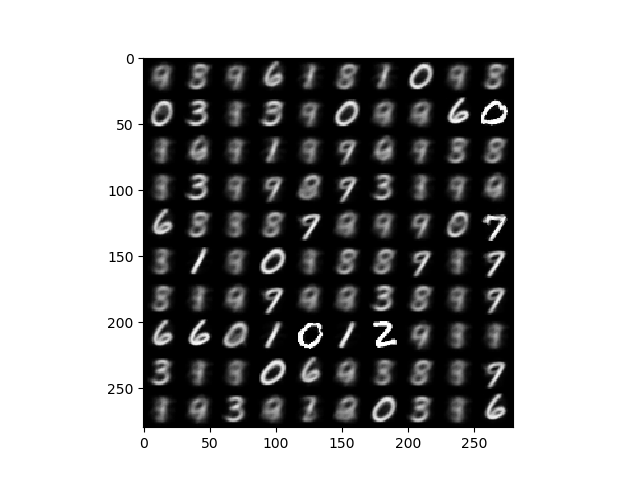}
        \caption{Reconstructed digits.}
        \label{fig:reduced-mnist:blir_vae_4bits_recon}
    \end{subfigure}
    \begin{subfigure}{0.49\columnwidth}
        \includegraphics[width=\textwidth]{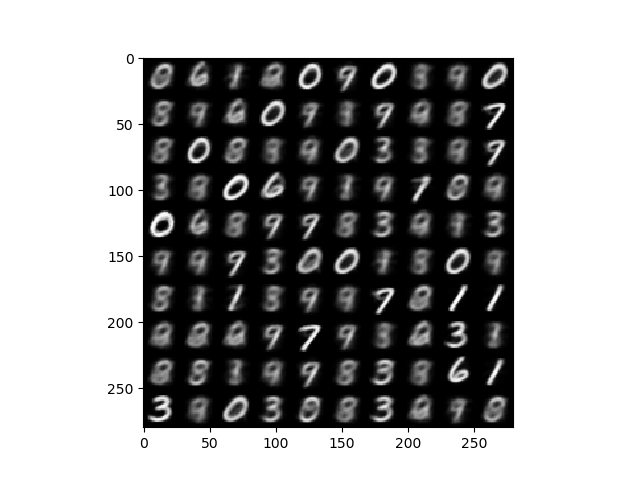}
        \caption{Generated digits.}
        \label{fig:reduced-mnist:blir_vae_4bits_gen}
    \end{subfigure}
    \caption{Figures taken from a BIR-VAE model trained on a 600 element subset of MNIST, the information rate is $\approx$ 2 bpi.}
\end{figure}

Figures \ref{fig:reduced-mnist:info_vae_2lf_recon} and
\ref{fig:reduced-mnist:blir_vae_4bits_recon} show the digit reconstructions for the
InfoVAE and BIR-VAE ($\approx$ 2 bpi) respectively. The InfoVAE reconstructions are
significantly sharper, but artefacts can be observed in the images. A similar
observation can be made in the generated samples, shown in figures
\ref{fig:reduced-mnist:info_vae_2lf_gen} and \ref{fig:reduced-mnist:blir_vae_4bits_gen}.  

Using the BIR-VAE allows the information rate of the encoding channel to be set judiciously. Restricting the information rate reduces the likelihood (increasing the error of the reconstructions), but the goal is to learn a good generative model as well as achieve good reconstructions. The InfoVAE produces crisp reconstructions and generations, whereas the BIR-VAE models produce images that are blurry. However, overall the quality of the $\approx$ 2 bpi BIR-VAE is better than the InfoVAE, confirming the results of table \ref{table:reduced-mnist-mse}. The BIR-VAE produces images with fewer artefacts.  

\subsubsection{Sharpness of Generated Digits}
The BIR-VAE results produce blurry reconstructions and generations at lower information
rates. This is natural given the usage of a likelihood measure in combination with the
  assumption of a Gaussian distribution at the output (which leads to a squared error
  criterion). Increasing the information rate of the encoder 
channel (if enough data is present) improves the sharpness of the resulting images. It is
to be expected that a higher dimensionality of the latent variable space performs better
for higher rates. In a space of higher dimensionality the range of a particular digit (or
subclass of a digit) has more neighboring digits (subclasses), facilitating re-arrangement
and, hence learning. In a five-dimensional space, it is possible to achieve an information
rate of $\approx$ 33 bpi with $\sigma_\epsilon=0.01$. We display only the generated digits and not
the reconstructions as the reconstructions are of higher quality.  

\begin{figure}
    \centering
    \begin{subfigure}{0.49\columnwidth}
        \includegraphics[width=\textwidth]{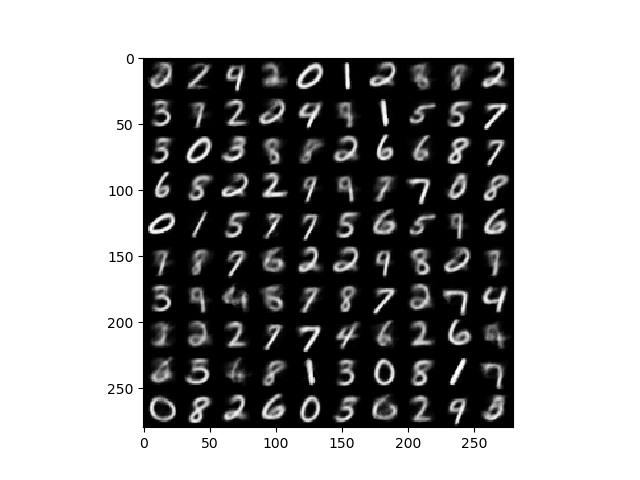}
        \caption{Two latent dimensions.}
        \label{fig:mnist:blir_vae_66bits_2lf_gen}
    \end{subfigure}
    \begin{subfigure}{0.49\columnwidth}
        \includegraphics[width=\textwidth]{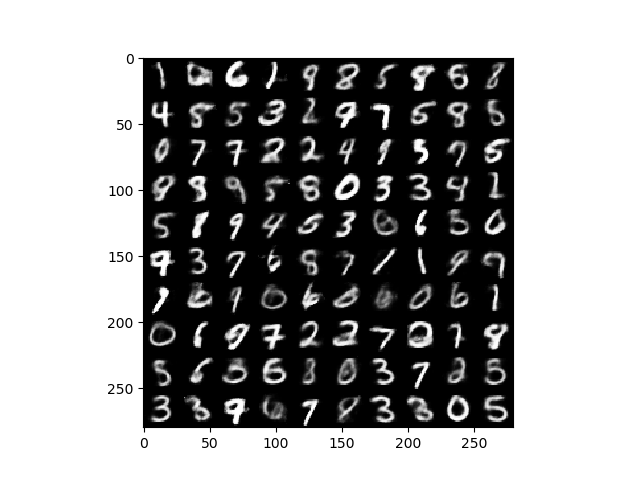}
        \caption{Five latent dimensions.}
        \label{fig:mnist:blir_vae_66bits_5lf_gen}
    \end{subfigure}
    \caption{Generated samples taken from the BIR-VAE model trained on the MNIST dataset
      with two and five latent dimensions. Both models have an information rate of $\approx$ 33 bpi.}
      \label{fig:mnist:sharp_comparason}
\end{figure}

Figure \ref{fig:mnist:sharp_comparason} shows the generated digits for a BIR-VAE with a rate of $\approx$ 33 bpi for the dimensionalities of two and five of the latent space.  It is seen that for the two-dimensional latent space (figure \ref{fig:mnist:blir_vae_66bits_2lf_gen}) the performance does not increase significantly over the $\approx$ 13.3 bpi case.  However, for the BIR-VAE with a five-dimensional latent space, shown in \ref{fig:mnist:blir_vae_66bits_5lf_gen}, the degree of sharpness is increased significantly. This indicates that the degrees of freedom in the model affects learning, and hence the generative model quality independently of the information rate. 

\subsubsection{Performance on Street View House Numbers}
\begin{figure}
	\centering
	\begin{subfigure}{0.49\columnwidth}
		\includegraphics[width=\textwidth]{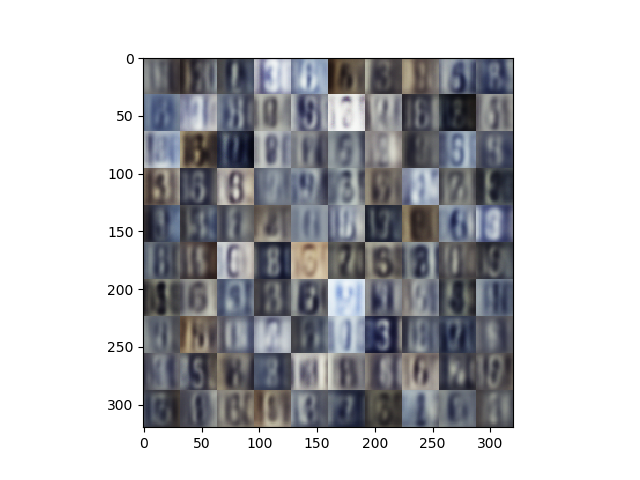}
		\caption{Rate: $\approx 66$ bpi.}
		\label{fig:svhn:blir_vae_132bits_20lf_gen}
	\end{subfigure}
	\begin{subfigure}{0.49\columnwidth}
		\includegraphics[width=\textwidth]{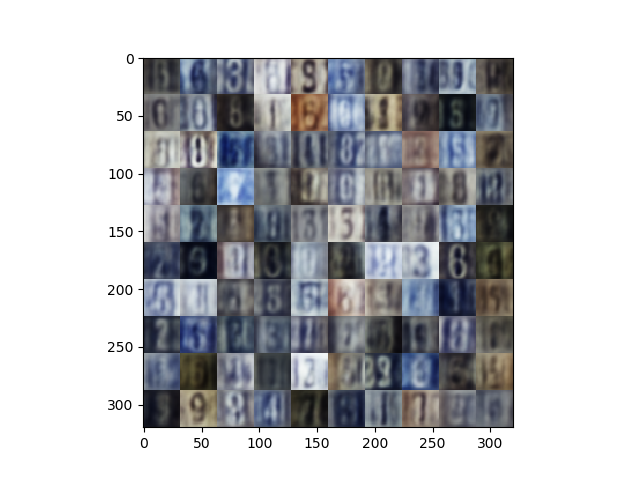}
		\caption{Rate: $\approx 132$ bpi.}
		\label{fig:svhn:blir_vae_265bits_20lf_gen}
	\end{subfigure}
	\caption{Generated samples taken from the BIR-VAE model trained on the SVHN dataset with 20 latent features and varying information rate.}
	\label{fig:bir_vae_svhn_comp}
\end{figure}

Figure \ref{fig:bir_vae_svhn_comp} compares the generated images from two BIR-VAEs on the
SVHN dataset, both with a 20-dimensional latent space.
The model with a higher information rate produces sharper and more convincing
results.

\begin{figure}
	\centering
	\begin{subfigure}{0.49\columnwidth}
		\includegraphics[width=\textwidth]{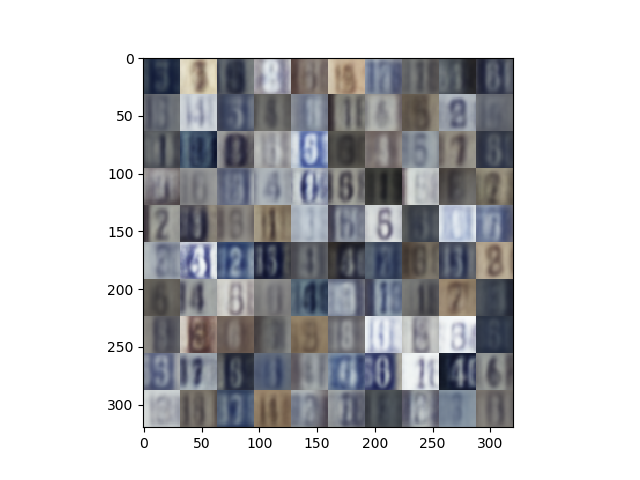}
		\caption{$\alpha = 0$.}
		\label{fig:svhn:info_vae_20lf_gen}
	\end{subfigure}
	\begin{subfigure}{0.49\columnwidth}
		\includegraphics[width=\textwidth]{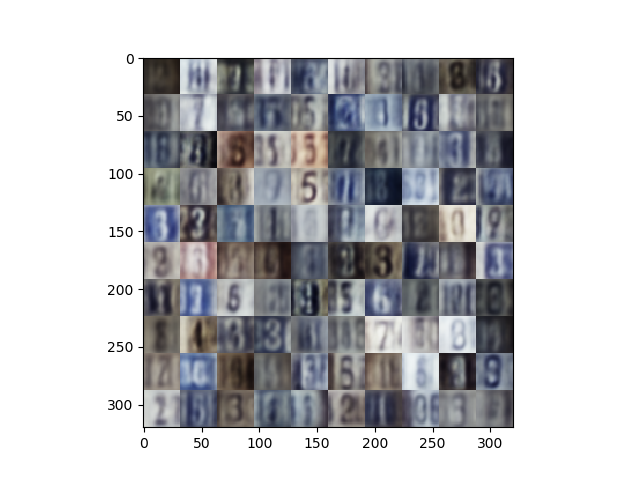}
		\caption{$\alpha = 0.9$.}
		\label{fig:svhn:info_vae_20lf_gen_ip}
	\end{subfigure}
	\caption{Generated samples taken from the InfoVAE model trained on the SVHN dataset with 20 latent features and varying information preference parameter, $\alpha$.}
	\label{fig:info_vae_svhn_comp}
\end{figure}

Figure \ref{fig:info_vae_svhn_comp} shows the InfoVAE model trained on the SVHN
dataset. When the information preference property of the the InfoVAE is set to $0$,
then the model 
generates simplistic samples which are not as detailed as either the BIR-VAE
models. In contrast, the InfoVAE with $\alpha = 0.9$ has a similar level of
generative quality as the BIR-VAE with an information rate of $\approx$ 132 bpi. 

\begin{table}
	\begin{tabular}{| c | c | c |}
		\hline
		Model & Train MSE & Test MSE \\
		\hline
		BIR-VAE ($\approx$ 66 bpi) & 16.72 & 17.97 \\
		BIR-VAE ($\approx$ 132 bpi) & 11.16 & 14.92 \\
		InfoVAE ($\alpha = 0$) & 17.32 & 20.50 \\
		InfoVAE ($\alpha = 0.9$) & 11.24 & 14.87 \\
		\hline
	\end{tabular}
	\caption{Mean Square Error (MSE) for the models trained on the SVHN dataset.}
	\label{table:svhn-mse}
\end{table}

Table \ref{table:svhn-mse} shows the MSE performance for each model trained on the SVHN dataset; comparing the performance of the two models further demonstrates that the InfoVAE with $\alpha = 0.9$ and the BIR-VAE with an information rate of $\approx$ 132 bpi have equivalent performance.

\subsection{Discussion of the Experimental Results}
This section has shown that the ability to set the information rate of the encoder channel allows the quality of the model to be controlled precisely. While it is possible to set a similar bound on the information rate of an InfoVAE, this was not proposed as part of the InfoVAE model. The InfoVAE paradigm also does not guarantee that it will use the available information. 

The BIR-VAE was shown to perform at least as well as the InfoVAE, with additional
ability to use the information rate to prevent over-fitting. To facilitate learning, and
to obtain sharply defined samples at high rates, the dimensionality of the BIR-VAE must be set
appropriately. 

\section{Conclusion}
\label{s:conclusion}

The Bounded Information Rate Variational Autoencoder (BIR-VAE) is a new method for learning generative models with meaningful latent representations. By restricting the information rate of the encoding channel, the generative capacity of the BIR-VAE is constrained in a principled way. An important attribute of BIR-VAE is that in situations with limited data, restricting the channel capacity of the BIR-VAE prevents the model from overfitting.

 The idea of using the mutual information between the input and the latent representation to learn meaningful representations has been used by other models, e.g. \cite{alemifixing, phuong2018the}. Our experimental results show that the performance of the BIR-VAE is at least as good as that of competing algorithms. However, in contrast to competing methods, the BLIR-VAE does not require the explicit approximation or evaluation of the mutual information, thus reducing the computational complexity of the training. 

The BIR-VAE paradigm is simple and intuitive. It trains an encoder-decoder network where the output of the encoder is subject to the addition of iid Gaussian noise with a fixed variance $\sigma_\epsilon^2$, and the input to the decoder is enforced to be unit Gaussian. The choice of $\sigma_\epsilon^2$ determines the information rate conveyed. To obtain a desired information rate $I$, the variance of the additive noise is set to $\sigma_\epsilon^2 = 4^{-\frac{I}{d}}$, where $d$ is the dimensionality of the latent variables.

While not discussed in detail, the additive noise channel in the BIR-VAE algorithm can be replaced by a generic vector quantiser with similar statistics of its quantisation noise. The resulting bitstream can be entropy-coded, to obtain a rate that closely approximates the set rate of BIR-VAE.  Thus, BIR-VAE can be used as a trainable encoder-decoder system for storage or transmission.

\medskip

\bibliographystyle{acm}
\bibliography{ms} 

\begin{thebibliography}{10}

\bibitem{li}
{GAN}s, mutual information, and possibly algorithm selection?
\newblock {\em http://www.yingzhenli.net/home/blog/?p=421\/}.

\bibitem{huszar_2016}
{InfoGAN}: using the variational bound on mutual information (twice).
\newblock {\em
  http://www.inference.vc/infogan-variational-bound-on-mutual-information-twice/\/}.

\bibitem{alemifixing}
{\sc Alemi, A.~A., Poole, B., Fischer, I., Dillon, J.~V., Saurous, R.~A., and
  Murphy, K.}
\newblock Fixing a broken {ELBO}.
\newblock In {\em Proceedings of the 35th International Conference on Machine
  Learning\/} (Stockholmsmässan, Stockholm Sweden, 10--15 Jul 2018), J.~Dy and
  A.~Krause, Eds., vol.~80 of {\em Proceedings of Machine Learning Research},
  PMLR, pp.~159--168.

\bibitem{arjovsky2017wasserstein}
{\sc Arjovsky, M., Chintala, S., and Bottou, L.}
\newblock {W}asserstein generative adversarial networks.
\newblock In {\em Proceedings of the 34th International Conference on Machine
  Learning\/} (International Convention Centre, Sydney, Australia, 06--11 Aug
  2017), D.~Precup and Y.~W. Teh, Eds., vol.~70 of {\em Proceedings of Machine
  Learning Research}, PMLR, pp.~214--223.

\bibitem{barber2003algorithm}
{\sc Barber, D., and Agakov, F.}
\newblock The {IM} algorithm: a variational approach to information
  maximization.
\newblock In {\em Advances in Neural Information Processing Systems 16\/}
  (2004), S.~Thrun, L.~K. Saul, and B.~Sch\"{o}lkopf, Eds., pp.~201--208.

\bibitem{bell1995information}
{\sc Bell, A.~J., and Sejnowski, T.~J.}
\newblock An information-maximization approach to blind separation and blind
  deconvolution.
\newblock {\em Neural computation 7}, 6 (1995), 1129--1159.

\bibitem{bengio2013representation}
{\sc Bengio, Y., Courville, A., and Vincent, P.}
\newblock Representation learning: {A} review and new perspectives.
\newblock {\em IEEE transactions on pattern analysis and machine intelligence
  35}, 8 (2013), 1798--1828.

\bibitem{bowman2015generating}
{\sc Bowman, S.~R., Vilnis, L., Vinyals, O., Dai, A.~M., Jozefowicz, R., and
  Bengio, S.}
\newblock Generating sentences from a continuous space.
\newblock In {\em Conference on Computational Natural Language Learning\/}
  (2016).

\bibitem{chen2016infogan}
{\sc Chen, X., Duan, Y., Houthooft, R., Schulman, J., Sutskever, I., and
  Abbeel, P.}
\newblock {InfoGAN}: Interpretable representation learning by information
  maximizing generative adversarial nets.
\newblock In {\em Advances in Neural Information Processing Systems 29\/}
  (2016), D.~D. Lee, M.~Sugiyama, U.~V. Luxburg, I.~Guyon, and R.~Garnett,
  Eds., pp.~2172--2180.

\bibitem{chen2016variational}
{\sc Chen, X., Kingma, D.~P., Salimans, T., Duan, Y., Dhariwal, P., Schulman,
  J., Sutskever, I., and Abbeel, P.}
\newblock Variational lossy autoencoder.
\newblock In {\em International Conference on Learning Representations\/}
  (2017).

\bibitem{cover2012elements}
{\sc Cover, T.~M., and Thomas, J.~A.}
\newblock {\em Elements of information theory}.
\newblock John Wiley \& Sons, 2012.

\bibitem{erez2005lattices}
{\sc Erez, U., Litsyn, S., and Zamir, R.}
\newblock Lattices which are good for (almost) everything.
\newblock {\em IEEE Transactions on Information Theory 51}, 10 (2005),
  3401--3416.

\bibitem{goodfellow2014generative}
{\sc Goodfellow, I., Pouget-Abadie, J., Mirza, M., Xu, B., Warde-Farley, D.,
  Ozair, S., Courville, A., and Bengio, Y.}
\newblock Generative {A}dversarial {N}ets.
\newblock In {\em Advances in Neural Information Processing Systems 27\/}
  (2014), Z.~Ghahramani, M.~Welling, C.~Cortes, N.~D. Lawrence, and K.~Q.
  Weinberger, Eds., pp.~2672--2680.

\bibitem{gray2012source}
{\sc Gray, R.~M.}
\newblock {\em Source coding theory}, vol.~83.
\newblock Springer Science \& Business Media, 2012.

\bibitem{gregor2015draw}
{\sc Gregor, K., Danihelka, I., Graves, A., Rezende, D.~J., and Wierstra, D.}
\newblock Draw: A recurrent neural network for image generation.
\newblock In {\em Proceedings of the 32nd International Conference on Machine
  Learning\/} (Lille, France, 07--09 Jul 2015), F.~Bach and D.~Blei, Eds.,
  vol.~37 of {\em Proceedings of Machine Learning Research}, PMLR,
  pp.~1462--1471.

\bibitem{gretton2012kernel}
{\sc Gretton, A., Borgwardt, K.~M., Rasch, M.~J., Sch{\"o}lkopf, B., and Smola,
  A.}
\newblock A kernel two-sample test.
\newblock {\em Journal of Machine Learning Research 13}, Mar (2012), 723--773.

\bibitem{gulrajani2017improved}
{\sc Gulrajani, I., Ahmed, F., Arjovsky, M., Dumoulin, V., and Courville,
  A.~C.}
\newblock Improved {T}raining of {W}asserstein {GANs}.
\newblock In {\em Advances in Neural Information Processing Systems 30\/}
  (2017), I.~Guyon, U.~V. Luxburg, S.~Bengio, H.~Wallach, R.~Fergus,
  S.~Vishwanathan, and R.~Garnett, Eds., pp.~5767--5777.

\bibitem{higgins2016beta}
{\sc Higgins, I., Matthey, L., Pal, A., Burgess, C., Glorot, X., Botvinick, M.,
  Mohamed, S., and Lerchner, A.}
\newblock {Beta-VAE}: Learning basic visual concepts with a constrained
  variational framework.
\newblock In {\em International Conference on Learning Representations\/}
  (2017).

\bibitem{isola2017image}
{\sc Isola, P., Zhu, J., Zhou, T., and Efros, A.~A.}
\newblock Image-to-image translation with conditional adversarial networks.
\newblock In {\em Conference on Computer Vision and Pattern Recognition\/}
  (2017).

\bibitem{kingma2013auto}
{\sc Kingma, D.~P., and Welling, M.}
\newblock Auto-encoding variational {Bayes}.
\newblock {\em arXiv preprint arXiv:1312.6114\/} (2013).

\bibitem{kulkarni2015deep}
{\sc Kulkarni, T.~D., Whitney, W.~F., Kohli, P., and Tenenbaum, J.}
\newblock Deep convolutional inverse graphics network.
\newblock In {\em Advances in Neural Information Processing Systems 28\/}
  (2015), C.~Cortes, N.~D. Lawrence, D.~D. Lee, M.~Sugiyama, and R.~Garnett,
  Eds., pp.~2539--2547.

\bibitem{lecun1998mnist}
{\sc LeCun, Y.}
\newblock The {MNIST} database of handwritten digits.
\newblock {\em http://yann. lecun. com/exdb/mnist/\/} (1998).

\bibitem{ledig2017photo}
{\sc Ledig, C., Theis, L., Husz{\'a}r, F., Caballero, J., Cunningham, A.,
  Acosta, A., Aitken, A., Tejani, A., Totz, J., Wang, Z., et~al.}
\newblock Photo-realistic single image super-resolution using a generative
  adversarial network.
\newblock In {\em Conference on Computer Vision and Pattern Recognition\/}
  (2017).

\bibitem{linsker1988self}
{\sc Linsker, R.}
\newblock Self-organization in a perceptual network.
\newblock {\em Computer 21}, 3 (1988), 105--117.

\bibitem{makhzani2015adversarial}
{\sc Makhzani, A., Shlens, J., Jaitly, N., Goodfellow, I., and Frey, B.}
\newblock Adversarial autoencoders.
\newblock In {\em International Conference on Learning Representations\/}
  (2016).

\bibitem{netzer2011reading}
{\sc Netzer, Y., Wang, T., Coates, A., Bissacco, A., Wu, B., and Ng, A.~Y.}
\newblock Reading digits in natural images with unsupervised feature learning.
\newblock In {\em Advances in Neural Information Processing Systems workshop on
  deep learning and unsupervised feature learning\/} (2011), vol.~2011, p.~5.

\bibitem{phuong2018the}
{\sc Phuong, M., Welling, M., Kushman, N., Tomioka, R., and Nowozin, S.}
\newblock {The} {Mutual} {Autoencoder}: Controlling information in latent code
  representations, 2018.

\bibitem{rezende2014stochastic}
{\sc Rezende, D.~J., Mohamed, S., and Wierstra, D.}
\newblock Stochastic backpropagation and approximate inference in deep
  generative models.
\newblock In {\em Proceedings of the 31st International Conference on Machine
  Learning\/} (Bejing, China, 22--24 Jun 2014), E.~P. Xing and T.~Jebara, Eds.,
  vol.~32 of {\em Proceedings of Machine Learning Research}, PMLR,
  pp.~1278--1286.

\bibitem{saxe2018information}
{\sc Saxe, A.~M., Bansal, Y., Dapello, J., Advani, N., Kolchinsky, A., Tracey,
  B.~D., and Cox, D.~D.}
\newblock On the information bottleneck theory of deep learning.
\newblock In {\em International Conference on Learning Representations\/}
  (2018).

\bibitem{shwartz2017opening}
{\sc Shwartz-Ziv, R., and Tishby, N.}
\newblock Opening the black box of deep neural networks via information.
\newblock {\em arXiv preprint arXiv:1703.00810\/} (2017).

\bibitem{tishby2000information}
{\sc Tishby, N., Pereira, F.~C., and Bialek, W.}
\newblock The information bottleneck method.
\newblock {\em arXiv preprint physics/0004057\/} (2000).

\bibitem{zhao2018infovae}
{\sc Zhao, S., Song, J., and Ermon, S.}
\newblock {InfoVAE}: Balancing learning and inference in variational
  autoencoders.
\newblock {\em arXiv preprint arXiv:1706.02262v3\/} (2018).

\bibitem{zhu2017unpaired}
{\sc Zhu, J., Park, T., Isola, P., and Efros, A.~A.}
\newblock Unpaired image-to-image translation using cycle-consistent
  adversarial networks.
\newblock In {\em International Conference on Computer Vision\/} (2017).

\end{thebibliography}

\end{document}